\documentclass[10pt,twocolumn,letterpaper]{article}

\usepackage{wacv}
\usepackage{times}
\usepackage{epsfig}
\usepackage{graphicx}
\usepackage{amsmath}
\usepackage{amssymb}
\usepackage{booktabs}
\usepackage{array}
\usepackage{xspace}
\usepackage{cleveref}
\usepackage{comment}
\usepackage{caption}
\usepackage[table,xcdraw]{xcolor}
\usepackage{tabularx}

\def\wacvPaperID{18} 

\wacvalgorithmstrack   

\wacvfinalcopy 

\begin{document}
\newcommand{\MethodName}{ReMark\xspace}

\title{\MethodName: Receptive Field based Spatial WaterMark Embedding Optimization using Deep Network}
\author{Natan Semyonov\\
\and
Rami Puzis\\
\and
Asaf Shabtai\\
\and
Gilad Katz\\
\and
Department of Software and Information Systems Engineering
\and
Ben Gurion University of the Negev\\
}

\maketitle
\thispagestyle{empty}

\begin{abstract}
Watermarking is one of the most important copyright protection tools for digital media.  
The most challenging type of watermarking is the imperceptible one, which embeds identifying information in the data while retaining the latter's original quality.
To fulfill its purpose, watermarks need to withstand various distortions whose goal is to damage their integrity. 
In this study, we investigate a novel deep learning-based architecture for embedding imperceptible watermarks. The key insight guiding our architecture design is the need to correlate the dimensions of our watermarks with the sizes of receptive fields (RF) of modules of our architecture. This adaptation makes our watermarks more robust, while also enabling us to generate them in a way that better maintains image quality.  
Extensive evaluations on a wide variety of distortions show that the proposed method is robust against most common distortions on watermarks including collusive distortion.
\end{abstract}

\section{Introduction}
Commercial over-the-top media services usually provide  subscription-based services offering online streaming of digital content. 
To defend against piracy, the providers embed additional information, referred to as transparent (imperceptible) watermark, into the digital media. While a watermarked image may appear normal to a casual observer, a knowledgeable recipients can extract the watermark to determine its source. A dishonest observer (attacker) looking to redistribute the content will attempt to corrupt the watermark to protect one's identity. The goal of the providers, therefore, is to make the watermark both difficult to detect and highly robust.

\begin{figure*}[t]
\centering
\includegraphics[width=1\linewidth]{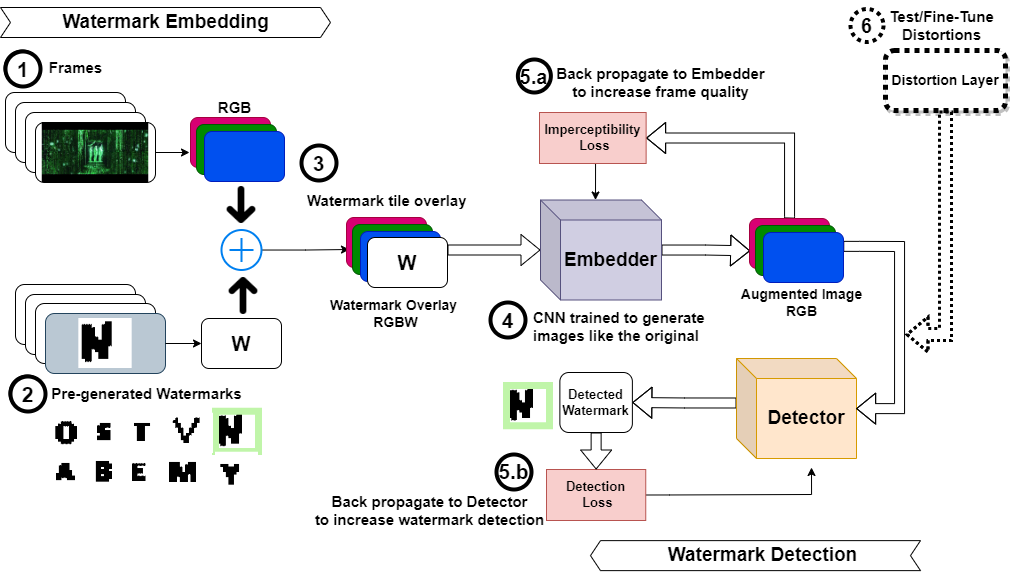}
\caption{The proposed \MethodName architecture for spatial watermark embedding. 
First, we split the video frame into its RGB layers.
Randomize watermark from a given set. Overlay it onto the chosen individual video frame by watermark tiling, resulting in a four-layer image (RGBW), called watermark overlay. 
Our embedder receives as input the 4-channel image and outputs a three-layer RGB watermarked image called augmented image.
The watermarked image is then fed to the Detector that detects the correct watermark.
Training is described in steps 1-5. 
Fine-Tuning is described in step 6.}
\label{fig:TrainingArchitecture}
\end{figure*}

In this study, we present Receptive waterMarking (\MethodName), a novel, highly robust, receptive field (RF)-based neural architecture for the embedding of spatial watermarks in images. 
In our proposed approach, shown in Figure \ref{fig:TrainingArchitecture}, a watermark is overlaid on top of the image as a new spatial data channel. Next, the Embedder network blends the watermark channel into the three original RGB channels, resulting in an augmented image. \MethodName then uses two loss functions---\textit{imperceptibility loss} and \textit{detection loss}---to balance maintaining image quality with the need to create robust watermarks.

\MethodName's watermark generation process is guided by the RF sizes (i.e., ``scope'' of vision) of its Embedder and Decoder components. First, by creating watermarks that are \textit{larger} than the Embedder's RF, we ensure that fragments of the watermark are spread across large swaths of the augmented image. Secondly, by ensuring the watermark is \textit{smaller} than the Detector's RF, we make the detection of the former easier and more robust.


The main contributions of this study are as follows: 
\begin{enumerate}
    \item A novel approach that utilizes the architecture's RF size for the generation of more robust watermarks.	
    \item The development of a watermark detector that is robust against multiple common distortions. Ours is also the first work to be evaluated on, and be robust against, collusion  attacks, which are used in video fraud.
    \item We present a neural architecture for the embedding of spatial watermarks that is sufficiently efficient to support high definition ($1280\times720$) images while utilizing a single standard GPU. 
\end{enumerate}

\section{\label{sec:related}Background and Related Work}

\subsection{Watermarking in Images and Video}
Effective watermarks need to be \textit{imperceptible}, \textit{detectable}, and \textit{robust}. Imperceptibility, which is critical so as not to interfere with viewer experience, is evaluated based on its similarity to the original image. The metrics used to evaluate this metric include peak signal to-noise ratio (PSNR)~\cite{5596999} and mean structural similarity (SSIM)~\cite{5596999}. Detectability refers to the ability to correctly identify the watermark in an embedded image, while robustness measures one's ability to detect the watermark after it has been distorted. Distortion techniques \cite{singh2013survey} include, but are not limited to, Jpeg compression, or collusion. In addition to these three metrics, studio operators also distinguish between \textit{blind} or \textit{non-blind} watermarks. A non-blind watermark requires both the original and watermarked content for its inference process \cite{650120,Hsieh2001HidingDW}, while blind watermarks require only the watermarked content. In this paper we focus on blind watermarking, which is the preferred modus operandi.

The field of watermarking has seen significant progress over the last decade. While earlier techniques sought to place the watermarks in the least significant (i.e., informative) regions of the image \cite{413536}, these watermarks were easy to detect and remove. More advances methods sought to encode the watermark in images' frequency representation \cite{Hsieh2001HidingDW,846253,1560462,4271513}. In these studies, values of certain frequencies are altered using common transformation methods such as discrete cosine transform~\cite{1672377}, discrete wavelet transform~\cite{1273384}, and discrete Fourier transform ~\cite{winograd1978computing}. Other methods combine frequency-domain encoding with log-polar mapping~\cite{1227605} or template matching~\cite{1560462} to achieve robustness against spatial domain transformations. 

The state-of-the-art performance achieved by deep neural networks in multiple vision-related tasks has led to their adaptation for watermarking tasks. HiDDeN~\cite{zhu2018hidden}, the first work to utilize deep architectures for this task, used an auto-encoder to embed a message within the cover image. RedMark ~\cite{AHMADI2020113157} extends the HiDDeN through the use of space-to-depth transforms and circular convolutions. StegaStamp ~\cite{DBLP:journals/corr/abs-1904-05343} focus on dealing with modelling more complex and realistic primitive transforms such as color-change via camera capture. Another work by Zhang et al.~\cite{zhang2019robust} introduces RIVAGAN, which features a custom attention-based mechanism. DVMark~\cite{DBLP:journals/corr/abs-2104-12734} is a recent study that employs a multi-scale design, where the watermark is distributed across multiple spatial-temporal scales.


\subsection{Receptive Fields in Vision-based Tasks}

Deep convolutional architectures employs layers of neurons that, using kernels, receive their inputs from similarly-located neurons in preceeding layers. 
When dealing with high-dimensional inputs such as images, it is computationally prohibitive to use dense architectures. Instead, we connect each neuron only to a local region in the input. The spatial extent of this connectivity is a hyperparameter called the receptive field of the neuron (equivalently this is the filter size). 
Therefore, the RF is defined as the size/range of the input which influences a particular neuron in a convolutional architecture, with this size being affected both by the convolutional and pooling layers.

\begin{figure}[h]
  \centering
   \includegraphics[width=0.7\linewidth]{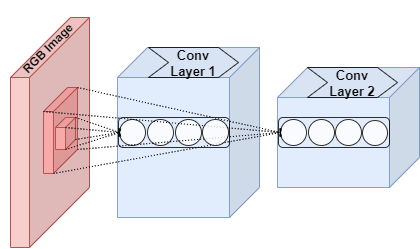}
   \caption{Illustration of RF. 
   Neurons of a convolutional layer (blue), connected to their RF (smaller red square).
   An input data (red) with an input shape of $width \times high \times channels$.
   The neurons in the second convolution has a bigger RF that the first convolution as it looks at bigger region in the input data.}
   \label{fig:LocalReceptiveField}
\end{figure}

By considering the receptive fields, we are able to ``assign'' each neuron to a specific region in the input data. This region is where the neuron will seek (and upon finding, will respond) to identifiable patterns in the data (e.g., lines, edges, textures). An illustration of the receptive field is presented in Figure~\ref{fig:LocalReceptiveField}.

The RFs of neurons are taken into consideration in multiple applications. In object detection tasks, for example, it is important to represent objects at multiple scales in order to recognize both small and large instances. Too small a RF may prevent the detection of large objects. \textit{It is one of the main hypotheses of this study that linking the size of the watermark with the size of \MethodName's architecture RFs will have significant impact on performance}.

Receptive fields play an important role in many state-of-the-art vision-related tasks such as object localization \cite{NIPS2013_f7cade80}, object detection \cite{Liu_2018_ECCV}, small object detection \cite{Du_2019} and image segmentation \cite{SULTANA2020106062}. For an in-depth analysis of the utilization of RFs, we refer reader to \cite{10.5555/3157382.3157645}.

Despite their diversity, no existing DL-based watermarking solution, to our knowledge, has taken the architecture's RFs into consideration. In this study, we explore the effects of aligning our generated watermarks with our architecture's RF, and demonstrate the efficacy of our approach.

\section{\label{sec:method}The Method}

\noindent \textbf{Overview.} Our proposed approach is presented in Figure~\ref{fig:TrainingArchitecture}. \MethodName consists of three components: \textit{Embedder}, \textit{Detector}, and \textit{Distortion Layer}. First, the original image (step \#1 in Figure~\ref{fig:TrainingArchitecture}) and the watermark (step \#2) are combined to create a joint representation (step \#3). This joint representation is fed to the Embedder (step \#4), which produces the augmented image. The goal of this image is to contain the watermark in its concealed form, while also not degrading the quality of the original image. 

The Embedder is trained in two stages. The first training stage consists of two loss functions: \textit{imperceptibility loss} and \textit{detection loss}. The goal of the former (step \#5.a) is to quantify the differences between the original and augmented image. The goal of the latter (step \#5.b) is to determine whether the watermark can still be detected by our Detector component. Similar to a GAN architecture, the imperceptibility loss is only used to update the Embedder, while the detection loss is used to update both the Embedder and the Decoder. The second training phase involves the Distortion Layer (step \#6), whose goal is to carry out various attacks on the augmented image in an attempt to corrupt the watermark. This training phase is used to make both the Embedder and the Detector more robust.

The remainder of this section is organized as follows: in Section \ref{subsec:trainingProcess} we describe the creation of the joint image and the first training phase (steps \#1-\#5). Next, in Section \ref{subsec:distortionLayer} we describe the second training phase and the types of attacks carried out by the distortion layer (step \#6). Section \ref{sec:receptive-fields} presents our approach for incorporating the RF into our watermark generation-detection process (particularly regarding watermark sizes). Finally, in Section \ref{subsec:architecture} we present a detailed description of our architecture components and their parameters.

\subsection{The First Training Phase} \label{subsec:trainingProcess}

We begin by sampling a frame from a video (step \#1), which we denote as the \textit{true image}. Next, we randomly sample a watermark from a large, pre-generated watermarks (step \#2). The chosen watermark, which is much smaller than the frame, is then tiled (i.e., replicated) so as to reach the dimensions of one of the channels of the original image. We then overlay the watermark over the true image, transforming a $\{H\times W\times3\}$ image to a $\{H\times W\times4\}$ one (step \#3). We refer to the new image as the \textit{watermark overlay}.

The watermark overlay is provided as input to the Embedder component, which produces the \textit{augmented image} as output (step \#4). The latter has the same dimensions as the true image, with the watermark (in the ideal use case) being imperceptibly and robustly concealed within the original channels. The architecture of the Embedder, as well as the other components, is described in detail in Section \ref{subsec:architecture}.

We now use two loss functions (step \#5) to fine-tune our Embedder and Detector components. The first function is the imperceptibility loss $\mathcal{L}_{imp}$ (step \#5a), which measures the pixel-wise difference between the original and augmented images. We use mean squared error as our metric: $\sum_{i=1}^{D}(x_i-y_i)^2$, where $x_i,y_j$ are pixels of the original and augmented images.

The second loss function, the detection loss $\mathcal{L}_{det}$ (step \#5b), measures the Detector's ability to detect the watermark in the augmented image. The output of the Detector is a softmax whose size is equal to the number of watermarks, and on this softmax we calculate multi-class cross-entropy: $-\sum_{c=1}^My_{o,c}\log(p_{o,c})$ where M is the number of classes, $y \in \{0,1\}$ indicates whether label c is correct, and p is the predicted probability.

Finally, the Detectors loss function, $\mathcal{L}_D$ is defined as $\mathcal{L}_D = \mathcal{L}_{det}$, while the Embedder's loss function $\mathcal{L}_{E}$ is defined as $\mathcal{L}_E = \gamma_1\mathcal{L}_{imp}+\gamma_2\mathcal{L}_{det}$, where $\gamma_1$ and $\gamma_2$ are configurable parameters. When used together, the two loss functions \MethodName to generate augmented images that are both highly similar to the true image, while also robustly concealing the watermark.

\subsection{The Second (Fine-Tuning) Training Phase}
\label{subsec:distortionLayer}
Following the first training phase, \MethodName is now able to effectively conceal a watermark within the augmented image. This training alone, however, does not prepare our approach to defend against the various methods used by attackers to corrupt the watermark. Therefore, we now perform an additional training of our architecture (both Embedder and Detector), in which the augmented image is modified by our \textit{Distortion Layer} prior to its analysis by the Detector. It is important to note that the training process is identical to the one described in Section \ref{subsec:trainingProcess}: both the Embedder and the Detector are updated using $\mathcal{L}_{imp}$ and $\mathcal{L}_{det}$. As a result, the Embedder will aim not only to create augmented images that conceal the watermark well, but also to conceal the watermark in a way that will be robust against distortions. The Detector will learn to also identify the watermarks in a perturbed state.

For each augmented image, the Distortion Layer (step \#6 in Figure \ref{fig:TrainingArchitecture}) randomly applies one of the following distortions:

\begin{itemize} 
    \setlength\itemsep{0em}
    \item  \textbf{Cropout} -- the image is cropped, then restored to its original resolution.
    \item \textbf{Dropout} -- randomly selected pixels are dropped and replaced with original image pixels.
    \item \textbf{Jpeg compression} -- a lossy form of compression, based on the discrete cosine transform.
    \item \textbf{Quantization} -- a lossy compression technique achieved by compressing a range of values in an image into a single quantum value.
    \item \textbf{Collusion} -- we assume the attacker has two copies of the image, each with a different watermark. The attacker then merges the copies in an attempt to corrupt the watermark. One watermark is the one \MethodName is tasked with detecting, while the other is randomly chosen from our pool. We evaluate two forms of merging: in one we average the values of each pixel, in the other we alternate between the two images, repeatedly selecting a single pixel from each.
   \item \textbf{Identity} -- the image is unchanged. The goal of this setting is to prevent \MethodName from ``forgetting'' how to process non-distorted images.
\end{itemize}

\noindent In our selection of distortion techniques, we closely follow the setup described in \cite{zhu2018hidden}, which is considered the current state-of-the-art (SOTA). The results of our evaluation, which compare \MethodName's performance to \cite{zhu2018hidden}, are presented in Section \ref{sec:relatedworkcompared}.

\subsection{Adapting Watermarks to Receptive Fields}
\label{sec:receptive-fields}
Our rationale in adapting the watermarks to consider the receptive fields is as follows: to prevent the watermark from being ``overrun'' by the Embedder, it is desirable that the size of the former be larger than the receptive field of a single neuron in the final layer of the latter. By doing so we ensure that the watermark, which we replicate and tile, affect all the neuron's in the Embedder. At the same time, our aim is that the watermark be smaller than the Detector's receptive field, so that every neuron will be able to view the watermark in its entirety. We hypothesize that these two setting will improve the robustness and detectability of our watermarks, respectively.

Based on this conjecture, we experiment with a watermark overlay by tiling multiple copies of the same watermark on the additional watermark channel. The dimensions of each tile size are determined by the RF sizes of our architectures.
Our experiments, presented in the following section, clearly show that taking the RF sizes into consideration has significant effect on the performance and robustness of our proposed approach.
To ensure that the tiled watermarks follow our desired rationale, we need to calculate the allowed size range of each individual watermark.

\begin{figure*}[h!]
    \centering
   \includegraphics[width=0.8\linewidth]{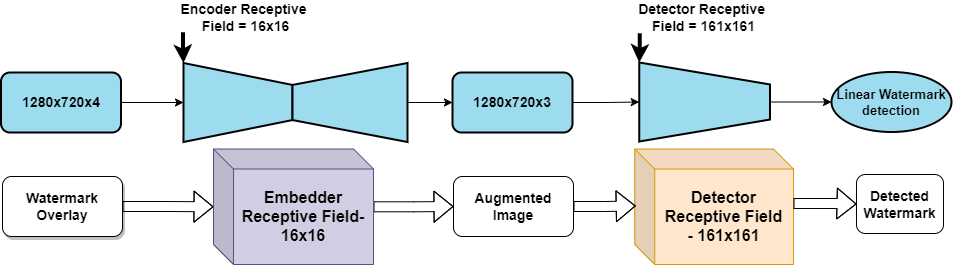}
   \caption{Receptive Field (RF) Flowchart of Embedder and Detector. The Embedder RF serves as a lower boundary of 16 for watermark size and the Detector RF as an upper boundary of 161.}
   \label{fig:ReceptiveField_ObjectRecognition}
\end{figure*} 

Our calculations of the watermark size are affected by the number of layers, stride size, kernel size, and padding. As shown in Figure~\ref{fig:ReceptiveField_ObjectRecognition}, the RF size for the Embedder is 16, and the Watermark Detector's is 161. 
As presented in Section~\ref{sec:receptiveFieldImpact}, watermark sizes outside this range suffer from significant performance degradation.

We now describe the method we used to calculate architecture RF.
We begin by calculating the number of output features, based on the number of input features and the properties of the employed convolutions. 
\begin{equation}
n_{out} = \left [ \frac{n_{in} + 2p - k}{s} \right ] + 1
\end{equation}
where:
\newenvironment{conditions}
  {\par\vspace{\abovedisplayskip}\noindent\begin{tabular}{>{$}l<{$} @{${}={}$} l}}
  {\end{tabular}\par\vspace{\belowdisplayskip}}
\begin{conditions}
 n_{in}    &  number of input features \\
 n_{out}     &  number of output features \\  
 k      & convolution kernel size \\
 p      & convolution padding size \\
 s      & convolution stride size  \\
\end{conditions}

Next we consider the jump in the output feature map, which is equal to the jump in the input map, multiplied by the number of input skipped features (i.e., the stride size).
\begin{equation}
j_{out} =   j_{in} * s
\end{equation}
where:
\begin{conditions}
 j_{in}    &  distance between two adjacent features into layer \\
 j_{out}     &  distance between two adjacent features out of layer  \\  
 s      & convolution stride size  \\
\end{conditions}

Finally, we calculate the size of the RF size of the output features map. This size is equal to the area covered by $k$ input features $[(k-1)*j_{in}]$, plus the extra area covered by the RF of the input feature.
\begin{equation}
r_{out} =   r_{in} + (k-1)*j_{in}
\end{equation}
where:
\begin{conditions}
 r_{in}    &  RF size of the input feature map \\
 r_{out}     &  RF size of the output feature map \\  
 j_{in}    &  distance between two adjacent features into layer \\
 k      & convolution kernel size \\
\end{conditions}

Using the above equations, we are able to calculate the valid watermark sizes ranges that satisfy our requirement for of our proposed architecture. 
The RF of the various layers of our architectures are presented in Table~\ref{tab:NetworksReceptiveField}.

\begin{table}[]
\centering
\resizebox{\columnwidth}{!}{%
\begin{tabular}{@{}lllll@{}}
Layer & Layer name & Map size (n\_in) & Jump (j\_in) & RF (r\_in) \\
0     & Input      & {[}1280, 720{]}  & 1.0          & 1.0                     \\
1     & Conv       & {[}320, 180{]}   & 4.0          & 4.0                     \\
2     & BatchNorm  & {[}320, 180{]}   & 4.0          & 4.0                     \\
3     & LeakyReLU  & {[}320, 180{]}   & 4.0          & 4.0                     \\
4     & Conv       & {[}160, 90{]}    & 8.0          & 16.0                    \\
5     & BatchNorm  & {[}160, 90{]}    & 8.0          & 16.0                    \\
6     & LeakyReLU  & {[}160, 90{]}    & 8.0          & {\textbf{16.0}}         \\
      &            &                  &              &                         \\
Layer & Layer name & Map size (n\_in) & Jump (j\_in) & RF (r\_in) \\
0     & Input      & {[}1280, 720{]}  & 1.0          & 1.0                     \\
1     & Conv       & {[}426, 240{]}   & 3.0          & 5.0                     \\
2     & ReLU       & {[}426, 240{]}   & 3.0          & 5.0                     \\
3     & MaxPool2d  & {[}141, 79{]}    & 9.0          & 17.0                    \\
4     & Conv       & {[}47, 26{]}     & 27.0         & 53.0                    \\
5     & ReLU       & {[}47, 26{]}     & 27.0         & 53.0                    \\
6     & MaxPool2d  & {[}15, 8{]}      & 81.0         & {\textbf{161.0}}   
\end{tabular}%
}
\caption{Embedder (upper) and Detector (lower) RF calculation. 
      The parameter n\textsubscript{in} calculates number of input features; j\textsubscript{in} calculates the distance between two adjacent features into layer; r\textsubscript{in} calculates the RF size of the input feature map. 
      RF of last layer and the entire network for Embedder and Detector is bold and underlined.}
\label{tab:NetworksReceptiveField}
\end{table}

\subsection{Architecture and Parameters} \label{subsec:architecture}
While both the Embedder and the Detector are convolution based architectures, the former is an autoencoder while the latter is a standard multi-label classifier.
A well-known problem in the application of convolutions and deconvolutions is the greater emphasis they place on specific areas of the image, due to uneven kernel overlap. 
A common result is an effect known as checkerboard artifacts, which greatly reduces image quality.
To avoid this phenomenon, we use Efficient Sub-Pixel Convolutional layer for Super-Resolution real-time single image and video content~\cite{shi2016real}.

The architecture of the Embedder is as follows: it receives as input a $(1280\times720\times4)$ image and outputs a $(1280\times720\times3)$ image. 
The Embedder consists of two 2D convolutional CONV units with leaky ReLUs, and batch normalization (BN) whose depths are respectively [16, 64]. 
These layers are followed by four transposed CONV units of depths [64,128,256, 1] respectively, each with PixelShuffle, leaky ReLU and BN, except for the final with a tanh. 
Our transposed CONV units have a stride of 1, and the kernel size is set to 3x3. We set the upscale of PixelShuffle to 2, the batch size to 6, and run the models for 400 epochs (each epoch does a cyclic pass).

The Watermark Detector receives as input a high definition image $(1280\times720\times3)$.
Its output is a softmax layer of the size equal to the number of possible watermarks. The architecture consists of two 2D convolutional CONV units, with ReLUs and MaxPool2d whose depths are respectively [16,32], followed by a multiclass-output fully connected (FC) unit with sigmoid activation. 

The learning rate used for both architectures was set to 0.0001, and the chosen optimization method was Adam. The Weights assigned to the detection and imperceptibility losses (see Section~\ref{subsec:trainingProcess}) were 0.05 and 0.95, respectively. These parameters were tuned by grid search to find the optimal balance for the performance of the Embedder and the Detector.

\begin{table*}[]
\centering
\resizebox{\textwidth}{!}{%
\begin{tabular}{@{}|l|l|c|c|c|c|c|c|c|@{}}
\toprule
\rowcolor[HTML]{F2F2F2} 
\textbf{Model type} &
  \textbf{Training Phase} &
  \multicolumn{1}{l|}{\cellcolor[HTML]{F2F2F2}\textbf{Cropout}} &
  \multicolumn{1}{l|}{\cellcolor[HTML]{F2F2F2}\textbf{Dropout}} &
  \multicolumn{1}{l|}{\cellcolor[HTML]{F2F2F2}\textbf{Jpeg compression}} &
  \multicolumn{1}{l|}{\cellcolor[HTML]{F2F2F2}\textbf{Quantization}} &
  \multicolumn{1}{l|}{\cellcolor[HTML]{F2F2F2}\textbf{Collusion (every second)}} &
  \multicolumn{1}{l|}{\cellcolor[HTML]{F2F2F2}\textbf{Collusion (average)}} 
  \\ \midrule
\rowcolor[HTML]{FFFFFF} 
R2D-16 {[}16x16{]} &
  Pre-training &
  \begin{tabular}[c]{@{}c@{}}66.8 \%\end{tabular} &
  \begin{tabular}[c]{@{}c@{}}65.6 \%\end{tabular} &
  \begin{tabular}[c]{@{}c@{}}\underline{20.8} \%\end{tabular} &
  \begin{tabular}[c]{@{}c@{}}71.5 \%\end{tabular} &
  \begin{tabular}[c]{@{}c@{}}\underline{44.4} \%\end{tabular} &
  \begin{tabular}[c]{@{}c@{}}68 \%\end{tabular} 
  \\ \midrule
\rowcolor[HTML]{FFFFFF} 
R2D-64 {[}64x60{]} &
  Pre-training &
  \begin{tabular}[c]{@{}c@{}}82 \%\end{tabular} &
  \textbf{\begin{tabular}[c]{@{}c@{}}99 \%\end{tabular}} &
  \textbf{\begin{tabular}[c]{@{}c@{}}\underline{67.6} \%\end{tabular}} &
  \textbf{\begin{tabular}[c]{@{}c@{}}99.7 \%\end{tabular}} &
  \textbf{\begin{tabular}[c]{@{}c@{}}99.4 \%\end{tabular}} &
  \textbf{\begin{tabular}[c]{@{}c@{}}99.1 \%\end{tabular}} 
  \\ \midrule
\rowcolor[HTML]{FFFFFF} 
R2D-1280 {[}1280x720{]} &
  Pre-training &
  \textbf{\begin{tabular}[c]{@{}c@{}}85.3 \%\end{tabular}} &
  \begin{tabular}[c]{@{}c@{}}80.2 \%\end{tabular} &
  \begin{tabular}[c]{@{}c@{}}\underline{18.5} \%\end{tabular} &
  \begin{tabular}[c]{@{}c@{}}99.6 \%\end{tabular} &
  \begin{tabular}[c]{@{}c@{}}\underline{27.9} \%\end{tabular} &
  \begin{tabular}[c]{@{}c@{}}98.3 \%\end{tabular} 
  \\ \midrule
\rowcolor[HTML]{F2F2F2} 
R2D-16 {[}16x16{]} &
  Fine-tuning &
  \begin{tabular}[c]{@{}c@{}}66.9 \%\end{tabular} &
  \begin{tabular}[c]{@{}c@{}}71.4 \%\end{tabular} &
  \begin{tabular}[c]{@{}c@{}}66.8 \%\end{tabular} &
  \begin{tabular}[c]{@{}c@{}}70.4 \%\end{tabular} &
  \begin{tabular}[c]{@{}c@{}}\underline{35.3} \%\end{tabular} &
  \begin{tabular}[c]{@{}c@{}}67.6 \%\end{tabular} 
  \\ \midrule
\rowcolor[HTML]{F2F2F2} 
R2D-64 {[}64x60{]} &
  Fine-tuning &
  \textbf{\begin{tabular}[c]{@{}c@{}}99.1 \%)\end{tabular}} &
  \textbf{\begin{tabular}[c]{@{}c@{}}99.5 \%\end{tabular}} &
  \textbf{\begin{tabular}[c]{@{}c@{}}99.7 \%\end{tabular}} &
  \textbf{\begin{tabular}[c]{@{}c@{}}99.9 \%\end{tabular}} &
  \textbf{\begin{tabular}[c]{@{}c@{}}99.5 \%\end{tabular}} &
  \textbf{\begin{tabular}[c]{@{}c@{}}99.7 \%\end{tabular}} 
  \\ \midrule
\rowcolor[HTML]{F2F2F2} 
R2D-1280 {[}1280x720{]} &
  Fine-tuning &
  \begin{tabular}[c]{@{}c@{}}98.1 \%\end{tabular} &
  \begin{tabular}[c]{@{}c@{}}99.1 \%\end{tabular} &
  \begin{tabular}[c]{@{}c@{}}99.6 \%\end{tabular} &
  \begin{tabular}[c]{@{}c@{}}99.8 \%\end{tabular} &
  \begin{tabular}[c]{@{}c@{}}\underline{93.2} \%\end{tabular} &
  \begin{tabular}[c]{@{}c@{}}99.3 \%\end{tabular} 
  \\ \bottomrule
\end{tabular}%
}
\caption{ 
       Watermark detection accuracy in presence of distortions described in Section \ref{subsec:distortionLayer} before and after fine-tuning. The highest the best performing models in each training phase are marked with bold. The most sever attacks for each model are marked with underline. } 
       
\label{tab:RobustnessTable}
\end{table*}

\section{Empirical Evaluation}

\subsection{Experimental Setup}
\textbf{Dataset:}
Preliminary experiments were performed with the CelebA dataset~\cite{liu2015faceattributes}. 
The final architecture, listed in Table~\ref{tab:NetworksReceptiveField}, was evaluated using a four-minute excerpt from the Matrix movie. 
The models were trained on 5,000 random HD quality frames from this excerpt. 
Please note that, although, generic watermarking models may save precious training time, in practice models may be fine-tuned for each individual movie. 
In order to show that the models need not be trained on all frames, we withhold 1000 frames that are used for evaluation.  
The test set includes withheld images distorted as described in Section~\ref{subsec:distortionLayer}.


\textbf{Watermarks:}
We used ten spatial watermarks, each watermark contains a single letter from the English alphabet.
We experiment with different sizes of the watermarks $[4\times 4,8\times 8,16\times 16,32\times 30,40\times 40 ,64\times 60,128\times 120,160\times 180, 320\times 360, 1280\times 720]$ chosen as round factors of the image size. 
The watermarks are tiled starting with the upper left corner of the image. 
We denote models trained with $16\times 16, 64\times 60$, and $1280\times 720$ watermarks as
\textbf{R2D-16}, \textbf{R2D-64}, and \textbf{R2D-1280} respectively. 

\textbf{Training hyper-parameters:}
We trained the models with batches of 6 images.
We ran a grid search on the learning rates of the optimizer and on the $\gamma_1$ and $\gamma_2$ factors of the $\mathcal{L}_{imp}$ and $\mathcal{L}_{det}$ losses.

\textbf{Performance metrics:}
To evaluate the models performance we measured: 
(1) the imperceptibility loss $\mathcal{L}_{imp}$, 
(2) PSNR and SSIM, 
(3) the detection loss $\mathcal{L}_{det}$,
and (4) the number of correctly detected watermarks out of the six images in each batch (denoted as Detection Accuracy).



Our evaluation is performed as follows. 
In Section~\ref{subsubsec:PreliminaryResults} we perform a preliminary analysis of our evaluated models, and compare their performance to the fine-tuning models for each of our evaluated distortions.
Next, in Section~\ref{sec:receptiveFieldImpact} we analyze the effect various watermark sizes have on the \MethodName's performance and demonstrating our hypothesis (watermark size between Embedder and Detector RFs achieves optimized performance). 
In Section~\ref{sec:distortion-training} we show the fine-tuning evaluation. \MethodName to withstand distortions and show favorable performance compared to a state-of-the-art architecture in Section~\ref{sec:relatedworkcompared} (Figure \ref{fig:relatedworkcompared}).  
Finally, in Section~\ref{sec:image-examples} we demonstrate high definition watermarked images.   


\subsection{Pre-training Results} \label{subsubsec:PreliminaryResults}

\begin{figure}[h]
  \centering
   \includegraphics[width=1\linewidth]{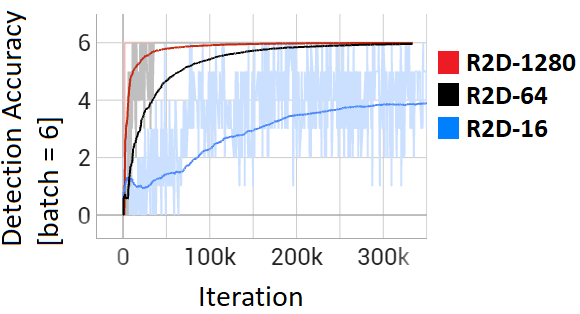}
   \caption{Detection Accuracy during pre-training (batch size of 6). One epoch includes 834 mini-batch iterations.}
   \label{fig:Training Phase}
\end{figure}

Figure~\ref{fig:Training Phase} presents the watermark detection accuracy during pre-training.  
R2D-1280 and R2D-64 eventually can perfectly detect watermarks on undistorted withheld images while R2D-16 fails to converge. 
Clearly, larger watermarks are easier to detect.

\begin{figure*}[h]
  \centering
   \includegraphics[width=0.7\linewidth]{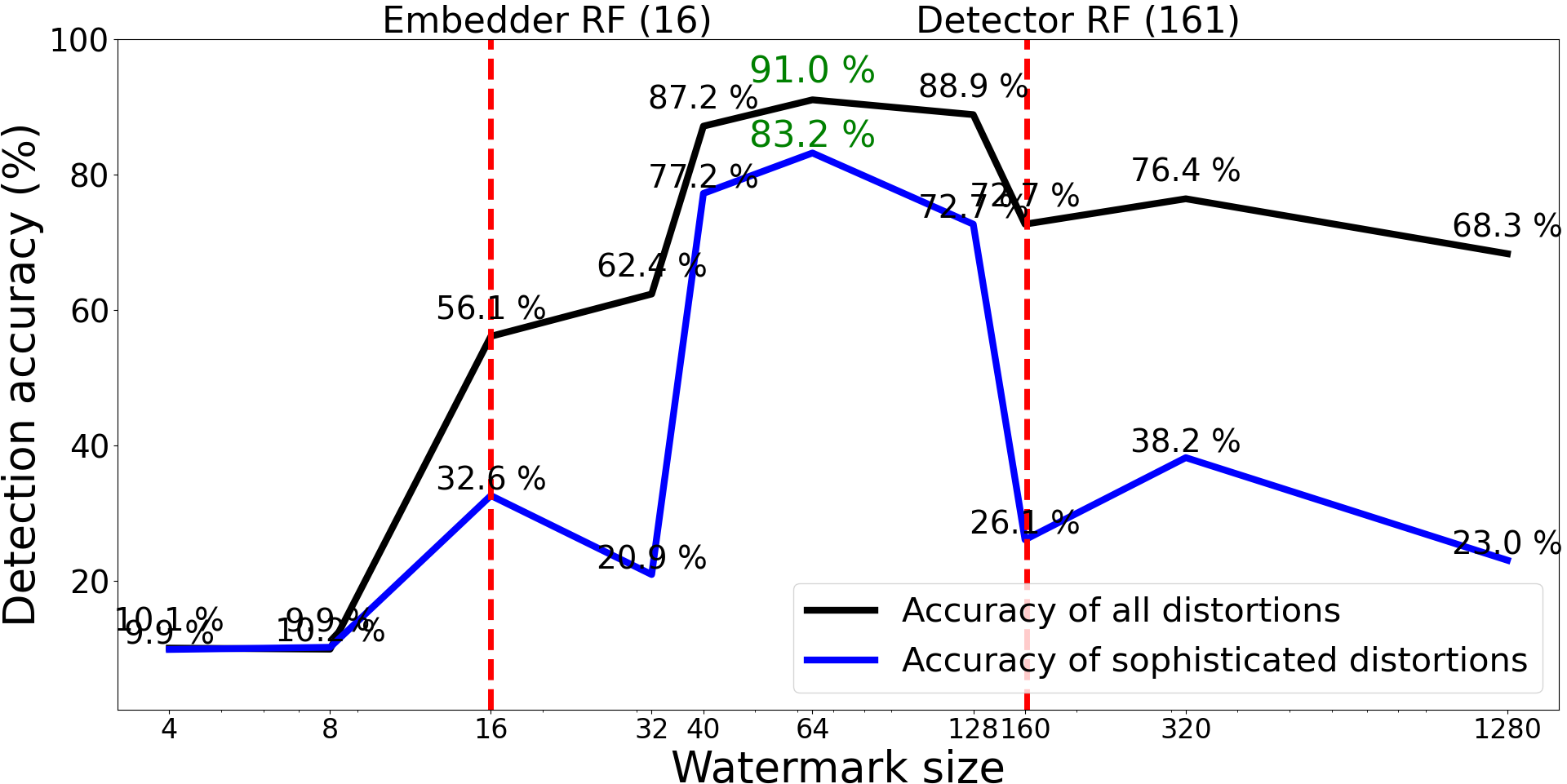}
   \caption{RF Boundaries affect on model robustness. 
   Each point in the graph denotes accuracy detection over different watermark sizes (4x4 - 1280x720). Black line describes average detection for all distortions. 
   Blue line describes average detection for most sophisticated distortions (Jpeg-Compression and Collusion (every second)). 
   Red dashed lines annotate the RF of embedder and detector networks.
   The best result is R2D-64 model (in Green) with watermark size of 64.}
   \label{fig:ReceptiveFieldAffect}
\end{figure*}

Next we evaluate the robustness of the pre-trained detectors to various distortions commonly used to tamper with watermarks.  
Table~\ref{tab:RobustnessTable} presents the watermark detection accuracy following the various distortions applied on the augmented images. 
Jpeg compression and Collusion (every second) cause the most severe degradation in the performance of all detectors (underlined in Table~\ref{tab:RobustnessTable}). 
We will refer to these two attacks as \emph{sophisticated distortions}.

Results indicate that R2D-64 is the most robust among the pre-trained models, even though the training curve of R2D-1280 is steeper. 
The only exception is cropout where R2D-1280 is slightly better. 
R2D-64 is robust against Dropout, Quantization, and Collusion even when trained with non-distorted images only. 
However, its accuracy degrades significantly following Cropout or Jpeg compression applied to the augmented images.  
We will show in Section~\ref{sec:distortion-training} how fine-tuning vis a vis image distortions mitigates the attacks. 
In conclusion, larger watermarks are easier to detect in mild conditions but they are not necessarily more robust against distortions.  
In the following subsection we investigate the effect of the watermark size on the performance of the pre-trained models with distortions.  


\subsection{Watermark Size and the Receptive Fields} \label{sec:receptiveFieldImpact}
Recall that the receptive fields of the Embedder and the Decoder are $16\times 16$ and $161\times 161$ respectively (see Table~\ref{tab:NetworksReceptiveField}).
These values are marked with red dashed lines in Figure~\ref{fig:ReceptiveFieldAffect}.
We evaluate the robustness of watermarks of various sizes ranging from $4\times 4$ to $1280\times 720$. 
The X-axis in Figure~\ref{fig:ReceptiveFieldAffect} represents the watermark width. 
The black line denotes the average detection rate for all distortions listed in Section~\ref{subsec:distortionLayer} and the blue line is the average detection accuracy for the most sophisticated distortions (Jpeg-Compression and Collusion-every second). 
Sophisticated distortions consistently cause higher reduction in the watermark detection accuracy. 

Next we focus on the effect of the watermark size on the detection accuracy. 
The highest detection accuracy is achieved for watermark sizes of 40, 64, and 128.  
These watermark sizes also exhibit lower performance difference between all and sophisticated distortions than 16, 32, 160, 320, and 1280. 
These results support our conjecture from Section~\ref{sec:receptive-fields} that the size of a detectable spatial watermark is constrained by the receptive fields of the Embedder and the Detector.    
In particular, the 64 watermark size, shown in green in Figure \ref{fig:ReceptiveFieldAffect}, exhibits the best performance in presence of distortions.


\subsection{Fine-tuning vis a vis Distortions}
\begin{figure}[h]
    \centering
    \includegraphics[width=1\linewidth]{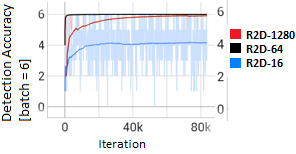}
    \caption{Fine-Tuning on Jpeg-Compression.}
    \label{fig:ReTrainAttacks}
\end{figure}

In this subsection we present the results of fine-tuning the Embedder and Detector models to the various distortions of the augmented images as described in Section~\ref{subsec:distortionLayer}.
During the training process the Distortion Layer was activated.
The rest of the training process is similar to pre-training.  
Figure \ref{fig:ReTrainAttacks} present the effects of the fine-tuning process on the watermark detection accuracy with Jpeg compression distortion. 
The zero's iteration corresponds to the pre-trained model. 
Surprisingly, unlike during pre-training, R2D-64 shows the fastest convergence. 
R2D-1280 eventually learns to withstand Jpeg compression but it takes more than 90 epochs (75K mini-batch iterations).   
R2D-16 after fine-tuning achieves the same performance as it showed with no distortions (see Figure~\ref{fig:Training Phase}).  

R2D-64 and R2D-1280 fine-tuned models exhibit improvement over their pre-trained versions. 
R2D-16 shows slight degradation against Quantization and Collusion attacks. 
R2D-64 shows the best overall performance improving from 67\% to 99\% on with Jpeg compression distortion.

Overall, R2D-64 adapts to distortions faster and better than 
R2D-16 with watermark size smaller than the Embedder RF and than R2D-1280 with watermark size larger than the Detector RF.
We note that this result can be attributed to the favorable performance of the pre-trained R2D-64 model.
Yet it stresses the importance of the careful choice of receptive filed and watermark sizes.


\label{sec:distortion-training}

\subsection{Comparison to State-of-the-Art} \label{sec:relatedworkcompared}
\begin{figure}[h]
    \centering
    \includegraphics[width=1\linewidth]{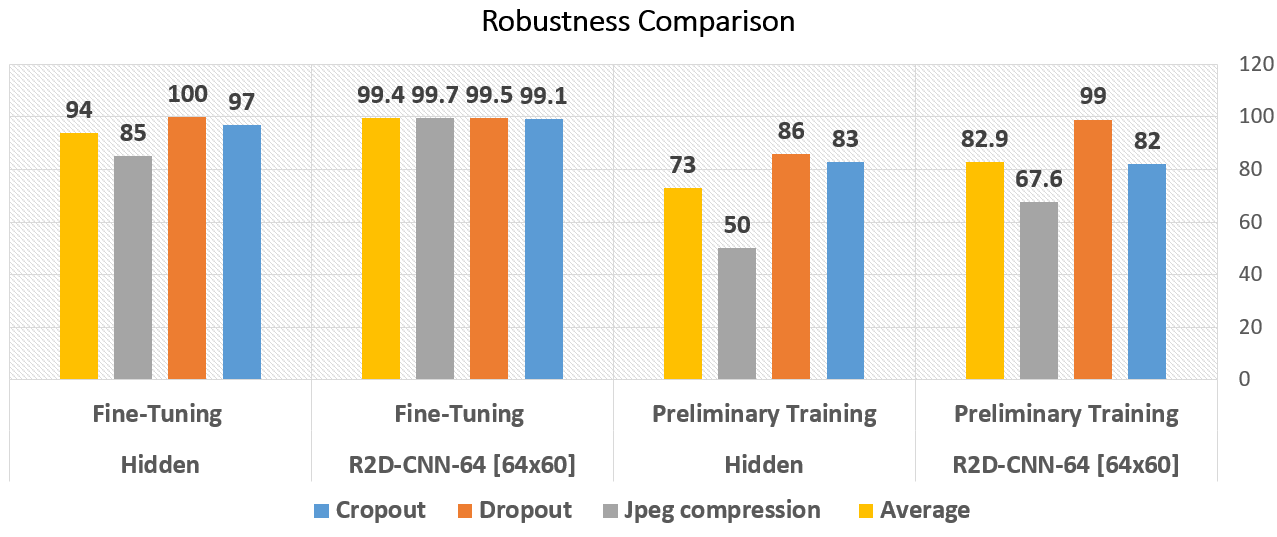}
    \caption{Robustness comparison. Accuracy per distortion of training and fine-tuning phases.}
    \label{fig:relatedworkcompared}
\end{figure}

In Figure~\ref{fig:relatedworkcompared} we compare the performance of R2D-64 to the state-of-the-art Hidden network~\cite{zhu2018hidden}. 
R2D-64 outperforms Hidden both after pre-training and after fine-tuning on all distortions evaluated by Zhu et al.~\cite{zhu2018hidden}.

Hidden and R2D-64 show similar trends across distortions with Droput being the easiest to withstand and Jpeg compression being the toughest. 
Fine-tuned R2D-64 manages to achieve better performance under Jpeg compression attack than under the other attacks.
Quantization and Collusion (every second) were not considered by Zhu et al. 
Finally, Hidden was trained and tested with $128\times 128$ images while we applied our models on HD images.

\subsection{HD Showcase}
\label{sec:image-examples}
\begin{figure}[h!]
    \centering
    \includegraphics[width=1\linewidth]{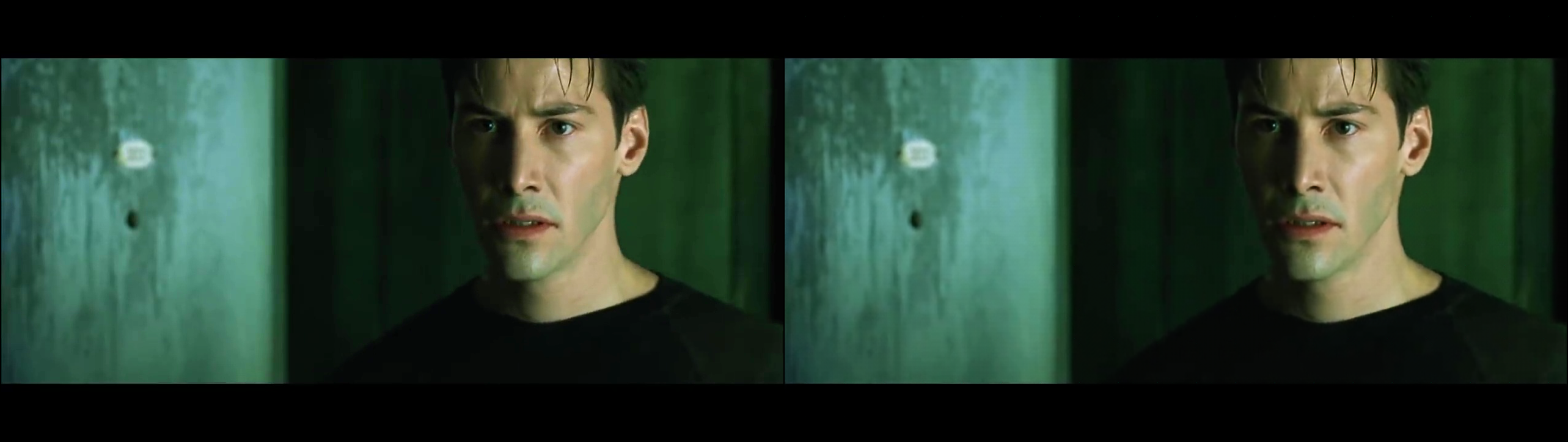}
    \includegraphics[width=1\linewidth]{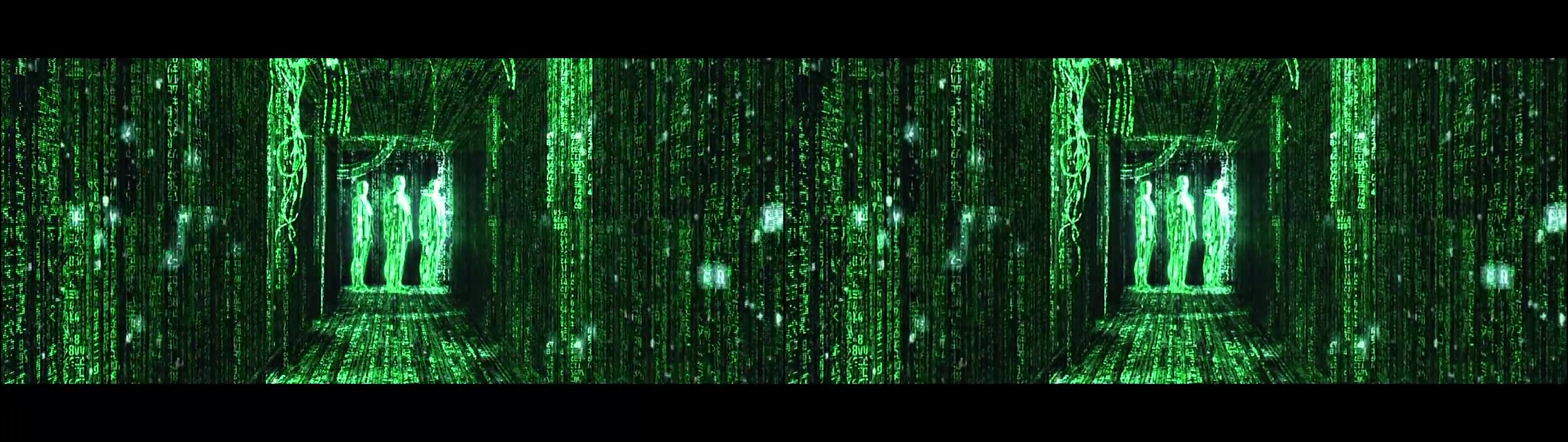}
    \caption{Original image (left) vs. Generated image (right). Zoom in by 575\% for HD quality on a Full HD display.}
    \label{fig:Images}
\end{figure}

Figure~\ref{fig:Images} demonstrates examples of the original frames from the Matrix movie vs. images augmented with imperceptible watermarks by the R2D-64 model. 
The reader is encouraged to zoom in to the images for a better view. 
In general we evaluated the quality of the augmented images with PSNR and SSIM. 
These metrics allow us to measure our method and test the quality and imperceptibility of watermark embedding. 
The average PSNR and SSIM values are 31.55  0.73, respectively. 

\section{Conclusions}
\label{sec:conclusions}

In this paper, we presented watermark embedding architecture optimization based on RF.
We developed a new end to end method to embed spatial watermarks into images. 
The developed method showed robustness and generalization on various distortions.

We take advantage of the RF in convolutional neural networks introducing an innovative approach for watermark embedding. 
To the best of our knowledge, this is the first application of RF for watermark embedding in convolutional architectures. 
The RF approach showed improved robustness compared to other watermark sizes that do not correspond with the RF boundaries of the architecture. 
Moreover, this approach shows high robustness even after pre-training without any knowledge on distortions. The moderated learning curve combined with high robustness signifies the generalization of the model.
In addition to robustness gained using the RF approach we are generating quality images that are indistinguishable from the original image.

Finally, collusion distortion firstly tested, which was not tested in related works.
Collusion is a very common attack in the video forgery. Here we present that our model is resistant to two alternatives of these distortions.  

For future work we believe that the RF based method can run on higher capacity of watermarks.

{\small
\bibliographystyle{ieee_fullname}
\bibliography{egbib}
}




\end{document}